\documentclass[journal]{IEEEtran}

\ifCLASSINFOpdf
\else
\fi

\usepackage{graphicx}
\usepackage{xcolor}
\usepackage{subcaption}
\usepackage{amsmath}
\usepackage{amssymb}
\usepackage{amsfonts}
\usepackage{tabularx}
\usepackage{makecell}
\usepackage{booktabs}
\usepackage{adjustbox}
\usepackage{multirow}
\begin{document}

\title{An Attention-based Framework with Multistation Information for Earthquake Early Warnings}

\author{Yu-Ming Huang, Kuan-Yu Chen,~\IEEEmembership{Member,~IEEE}, Wen-Wei Lin, and Da-Yi Chen

\thanks{Yu-Ming Huang, Kuan-Yu Chen, and Wen-Wei Lin are with the Department of Computer Science and Information Engineering, National Taiwan University of Science and Technology, Taipei 106, Taiwan (e-mail: sora@nlp.csie.ntust.edu.tw; kychen@mail.ntust.edu.tw; wenwei@nlp.csie.ntust.edu.tw).

Da-Yi Chen is with the Seismological Center, Central Weather Administration, Taipei 100, Taiwan (e-mail: dayi@cwb.gov.tw).}

\thanks{Manuscript received March XX, 2024; revised XXX XX, 2024.}}

\markboth{Journal of \LaTeX\ Class Files,~Vol.~13, No.~9, September~2024}%
{Shell \MakeLowercase{\textit{et al.}}: Bare Demo of IEEEtran.cls for Journals}

\maketitle

\begin{abstract}
Earthquake early warning systems play crucial roles in reducing the risk of seismic disasters. Previously, the dominant modeling system was the single-station models. Such models digest signal data received at a given station and predict earthquake parameters, such as the p-phase arrival time, intensity, and magnitude at that location. Various methods have demonstrated adequate performance. However, most of these methods present the challenges of the difficulty of speeding up the alarm time, providing early warning for distant areas, and considering global information to enhance performance. Recently, deep learning has significantly impacted many fields, including seismology. Thus, this paper proposes a deep learning-based framework, called SENSE, for the intensity prediction task of earthquake early warning systems. To explicitly consider global information from a regional or national perspective, the input to SENSE comprises statistics from a set of stations in a given region or country. The SENSE model is designed to learn the relationships among the set of input stations and the locality-specific characteristics of each station. Thus, SENSE is not only expected to provide more reliable forecasts by considering multistation data but also has the ability to provide early warnings to distant areas that have not yet received signals. This study conducted extensive experiments on datasets from Taiwan and Japan. The results revealed that SENSE can deliver competitive or even better performances compared with other state-of-the-art methods.

\end{abstract}


\IEEEpeerreviewmaketitle

\section{Introduction}
\IEEEPARstart{E}arthquake early warning systems are designed to issue early warnings within seconds after the onset of an earthquake to reduce the risk of earthquake disasters \cite{stidham1999three, gasparini2007earthquake, hsiao2009development, allen2009status,allen2019earthquake, wu2021earthquake}. These systems identify the seismic information of an earthquake based on signals received from stations and the differences in the propagation speeds of these statistics. When a seismic event occurs, stations sequentially detect seismic signals based on their distance from the epicenter. The information is transmitted to the earthquake early warning system to estimate the parameters of the earthquake, including information on the epicenter, intensity, and magnitude. Once the warning is confirmed, the system rapidly issues warnings to the areas expected to be affected through channels, such as text messages and television broadcasts.

Deep learning has recently demonstrated outstanding potential in various fields. Thus, deep learning-based methods, whose model structures are neural networks, have been proposed in the seismology context, including earthquake early warning \cite{cremen2020earthquake}, intensity prediction \cite{jozinovic2020rapid}, and p-phase picking \cite{9882131}. Most of these methods are trained using large amounts of seismic data and attributes, enabling them to understand the relationship between the received data and events \cite{cauzzi2016earthquake, kong2019machine, xie2020promise, chiang2022neural}. Most of the approaches belong to single-station modeling, which digests signal data received at a given station and predicts if an earthquake will occur at that location. Although the performances of these methods are acceptable and they rapidly deliver responses, they present certain inherent limitations. First, single-station models only consider data from a single seismic station as input while ignoring the rich information from other stations. Second, single-station models typically only alert local regions, lacking the ability to signal distant areas \cite{Cofré2022, song2022site}.

Owing to these challenges, multistation modeling, which considers information from multiple stations, has emerged as a suitable alternative \cite{munchmeyer2021transformer, ahn2023stable}. By leveraging information from numerous seismic stations, multistation modeling can deliver more comprehensive and accurate predictions of seismic events to enhance earthquake early warning performance. Previous studies have attempted to employ convolutional neural networks \cite{jozinovic2020rapid, jozinovic2022transfer}, graph convolutional neural networks \cite{kim2021graph}, or Transformer models \cite{munchmeyer2021transformer} to process information from multiple stations. These methods highlight the importance and effectiveness of multistation information for earthquake alarming.

Thus, we attempted to create a deep learning-based modeling system with multistation information for earthquake early warning. The motivation is twofold. First, we intended to investigate the feasibility, effectiveness, and efficiency of the currently used self-attention mechanism to model the relationships among a set of stations \cite{vaswani2017attention}. Second, recent methods only take the waveform, position, or a combination of both as station input. However, certain physical or geological information regarding each station can be used to further enhance the performance. Consequently, this study developed a framework called SENSE, which leverages the self-attention mechanism to consider multistation data and employs locality-specific embeddings to enable the model to learn and encode the specific characteristics of each station. Thus, SENSE is expected to produce more accurate predictions. We evaluated SENSE using the Japan and Taiwan earthquake datasets \cite{okada2004recent}. The present experiments showed that SENSE can achieve competitive or even better results than state-of-the-art baselines.

\section{Related Work}
\subsection{Transformer and Conformer Models}
\begin{figure*}
\begin{subfigure}{.33\textwidth}
  \centering
  \includegraphics[width=1.0\linewidth]{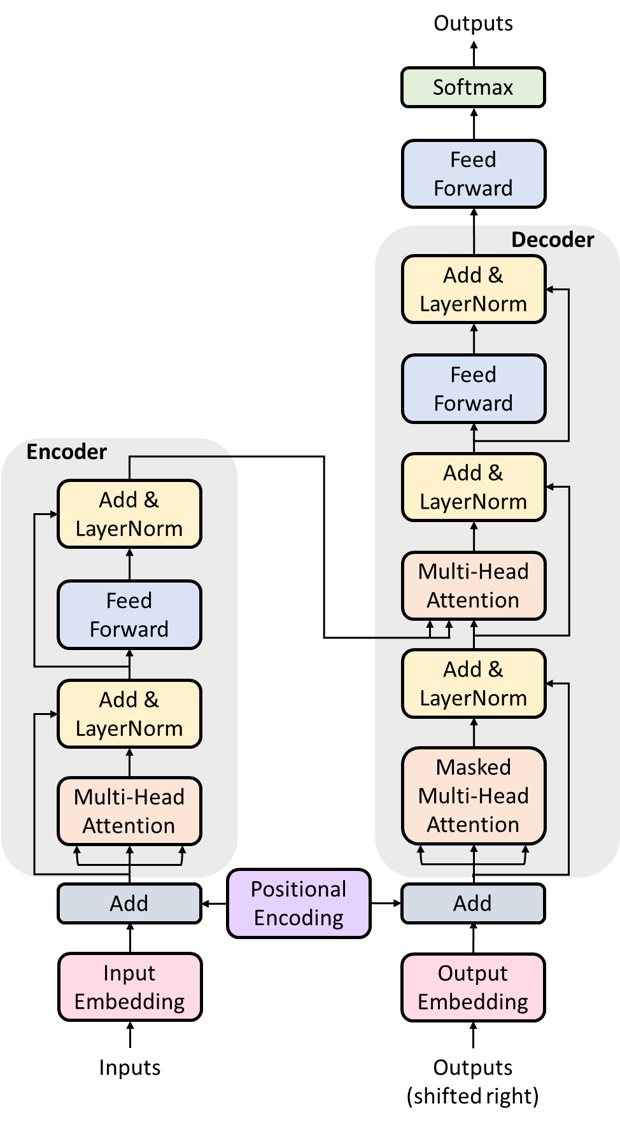}
  \caption{}
  \label{fig:Transformer}
\end{subfigure}%
\begin{subfigure}{.33\textwidth}
  \centering
  \includegraphics[width=.5\linewidth]{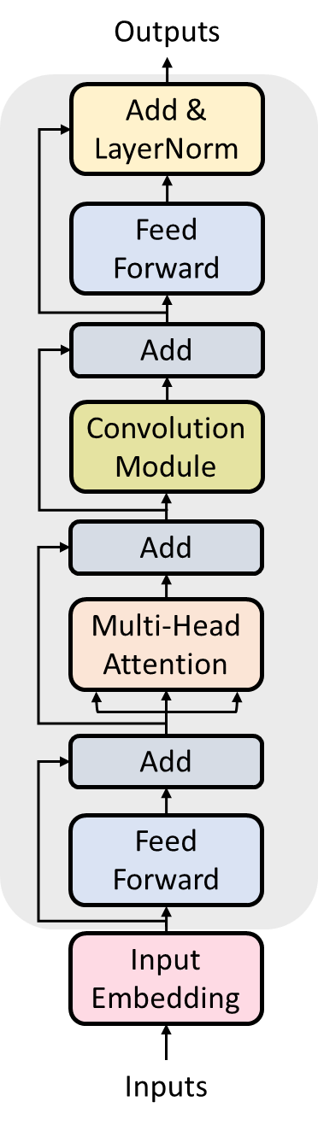}
  \caption{}
  \label{fig:Conformer}
\end{subfigure}
\begin{subfigure}{.33\textwidth}
  \centering
  \includegraphics[width=1.0\linewidth]{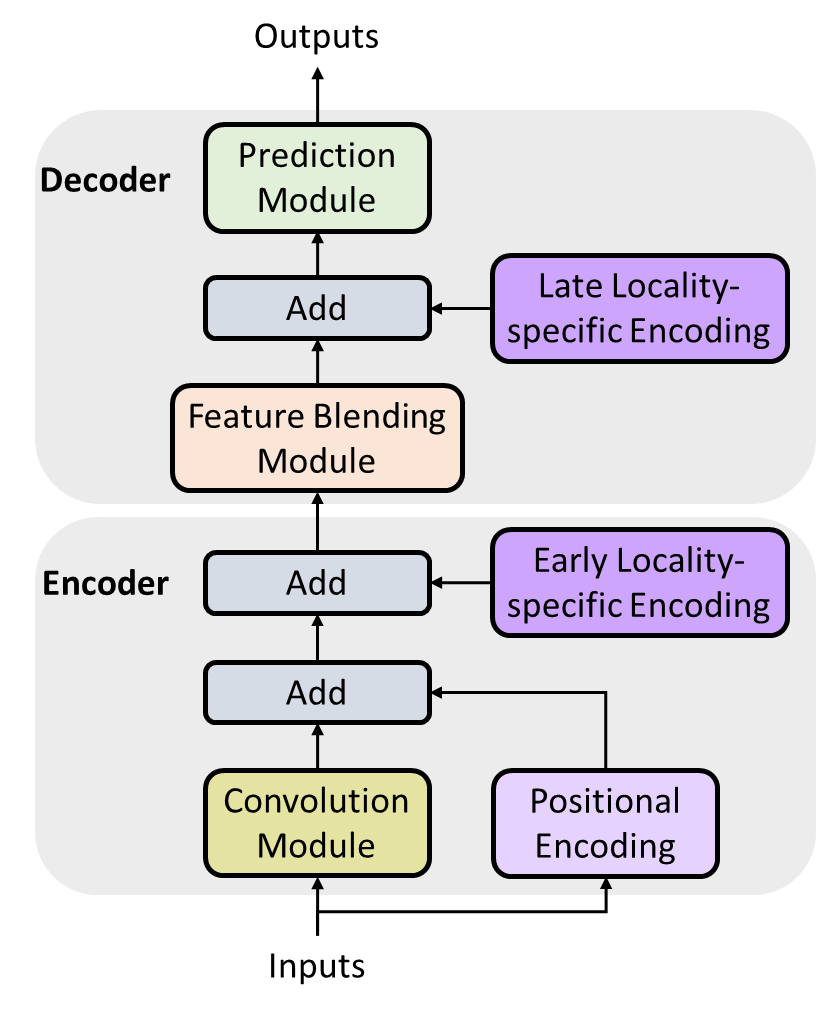}
  \caption{}
  \label{fig:SENSE}
\end{subfigure}
\caption{Model architectures of (a) Transformers, (b) Conformers, and (c) the proposed SENSE.}
\label{fig:Model}
\end{figure*}

The Transformer model has recently become a cutting-edge deep learning architecture in various fields \cite{vaswani2017attention}, such as natural language \cite{9894421}, speech \cite{9755057}, and vision processing \cite{dosovitskiy2020image}. Although the Transformer model was originally intended to handle time-series data and sequence generation problems, its ability to tackle common tasks was subsequently investigated and verified. The Transformer model exhibits an encoder–decoder architecture. The encoder dissects the input sequence while the decoder generates output sequences, rendering it an integral part of sequence-to-sequence tasks. Noteworthily, the encoder and decoder can be used individually or combined. Nevertheless, the key components of the encoder and decoder are the attention mechanism, positional encoding, feedforward neural network, layer normalization, and residual connection. Fig. \ref{fig:Model}(\subref{fig:Transformer}) shows the model architecture. Among the components, the fundamental innovation is the self-attention mechanism, which was leveraged in sequential modeling to process sequential data without relying on recurrence-based neural networks (i.e., vanilla recurrent neural networks \cite{lecun2015deep} and long short-term memory models \cite{hochreiter1997long}) or convolutional layers \cite{gu2018recent}. Dissimilar to the recurrence-based and convolutional models that sequentially process sequences, the self-attention mechanism enables the model to dynamically weigh the importance of each unit in a sequence relative to other units, capturing dependencies and relationships in a parallelized manner. The design allows the model to grasp long-range dependencies effectively. In addition to theoretical improvements, such a network architecture can easily handle large-scale datasets and speed up training because of its parallelizable design \cite{vaswani2017attention}.

Based on the fundamentals of the Transformer model, the Conformer model features pivotal modifications to enhance the efficiency and effectiveness of the Transformer encoder architecture \cite{tous2020deep}. The major components of a Conformer include the self-attention mechanism, convolution module, feedforward neural network, layer normalization, and residual connection. These components synergistically enable the model to discern intricate patterns, abstract representations, and hierarchical structures in a time-series data sequence. Fig. \ref{fig:Model}(\subref{fig:Conformer}) shows the model architecture. The main innovation is in the convolution module, which incorporates depthwise separable convolutions alongside the self-attention mechanism. The specialized design simultaneously performs convolutions across the time (sequence length) and depth (embedding dimensions) axes. Consequently, Conformers can capture long-term dependency, similar to the Transformers, and discern fine-grained local patterns.

Beyond their roots in natural language processing, the Transformer and Conformer models are versatile across domains, such as speech recognition, audio processing, and sequential data analysis \cite{guo2021recent,ma2021end,fang2021clip2video}. Their ability to capture dependencies across varying scales while maintaining computational efficiency renders them exceptional in diverse machine learning applications. Various studies have shown that they are promising frameworks for a wide range of real-world applications that rely on comprehensive sequence understanding and analysis.

\subsection{Deep Learning for Seismic Research}
Deep learning has recently become a cornerstone of seismological research. Through deep learning, many tasks have been improved, including earthquake intensity prediction \cite{song2022site,lomax2019investigation}, earthquake epicenter estimation \cite{kriegerowski2019deep, mousavi2019bayesian}, phase picking \cite{mousavi2020earthquake}, and earthquake early warning \cite{munchmeyer2021transformer,hong2021seismic}. Among popular neural networks, convolutional neural networks (CNNs) have been widely used in seismic data preprocessing, feature extraction and as building blocks for models \cite{o2015introduction, chiang2022neural}. Thibaut et al. \cite{perol2018convolutional} stacked multiple layers of CNNs for earthquake detection and source area estimation; they accurately predicted earthquake amplitudes. Hong et al. \cite{hong2021seismic} proposed two CNN-based models for estimating the surface response of earthquakes. Chiang et al. \cite{chiang2022neural} developed a CNN-based model to predict whether the peak ground acceleration (PGA) of a specific seismic monitoring station exceeded a predefined threshold or not. In addition to CNNs, recurrent neural networks (RNNs) are suitable building blocks for mitigating seismic tasks since the signal data from seismometers are time series. Chin et al. \cite{chin2020} employed multilayers of long short-term memory models (LSTMs) to identify the occurrence of an earthquake event and the durations of P- and S-waves. Berhich et al. \cite{Berhich2020} also used LSTM to create an earthquake prediction model. Cofré et al. \cite{Cofré2022} developed an LSTM-based method to address the problem of earthquake magnitude estimation for earthquake early warning. Berhich et al. \cite{BERHICH2022} examined the abilities of the LSTM and gated recurrent unit (GRU) to predict earthquake magnitude. To further combine the advantages of CNN and LSTM, Kail et al. \cite{Kail2022} proposed a recurrent CNN-based method to predict the location of earthquakes. 

Recently, neural networks based on attention mechanisms, particularly self-attention techniques, have become a mainstream architecture. The most famous representative is the Transformer \cite{vaswani2017attention}, which holds significant advantages over CNNs and RNNs in dealing with time-series data. Compared with RNNs, Transformers are parallelizable and can model long-term dependencies between data. Thus, they are more computationally efficient than RNNs. CNNs, although powerful for spatial data such as images, are not inherently designed for sequential data. Thus, they cannot effectively capture long-range context dependencies embedded in sequences. Owing to these advantages, studies have applied Transformers to seismic research. Mousavi et al. \cite{mousavi2020earthquake} proposed a very deep network architecture based on CNNs and Transformers for simultaneous earthquake detection and phase picking. Chin et al. \cite{chin2021attention} presented a Transformer-based hypocenter estimator for earthquake localization. Münchmeyer et al. \cite{munchmeyer2021transformer, Münchmeyer2020, Münchmeyer2020b, Münchmeyer2021b} used a series of Transformers to create an earthquake early warning system. Saad et al. \cite{saad2022real} employed the model architecture used in vision computing, which is also based on Transformers, to perform real-time earthquake detection and magnitude estimation. Although the Conformer is an extension that injects CNN components into the Transformer to obtain the best of both worlds, there is still a dearth of studies using the architecture in the context of seismological research.

\section{The Proposed Methodology}
With the rapid development of deep learning and the use of neural networks to enhance the performance of various seismic tasks, many interesting methods have been proposed. However, numerous existing methods only concentrate on a given seismic station or location. Therefore, these single-station methods are employed to process the data received from a station and deliver corresponding predictions regarding the location. To enhance the performance of an earthquake early warning system by leveraging information from multiple stations and advanced deep learning models, we present an attention-based framework called SENSE.

\subsection{Model Architecture}
SENSE is designed with an encoder–decoder architecture, as depicted in  Fig. \ref{fig:Model}(\subref{fig:SENSE}). The inputs comprise statistics from a set of stations in a given region or country. The output comprises intensity predictions for locations corresponding to the input stations. The encoder consists of a convolution module, a positional encoding module, and an early locality-specific encoding module. The decoder is based on a feature blending module, followed by a late local-specific encoding module and prediction module. Depending on the modeling strategy, even if only one station receives earthquake statistics because of the initial onset of an earthquake, SENSE is expected to signal early warnings to all locations related to the input stations.

The waveform $w\in{\mathbb{R}}^{3\times{T}}$ (i.e., three-axis acceleration of time $T$) and geographical information $g\in{\mathbb{R}}^{3}$ (i.e., longitude, latitude, and instrument height) are basic features of each station. For a given set of $N$ stations, $\{w_1,...,w_N\}$ and $\{g_1,...,g_N\}$ are input statistics to the model. The convolution module is initially employed to scan the waveform and encapsulate the information into feature vectors:
\begin{equation} 
    \label{WaveEmbedding_en}
    [W_1,...,W_N] = \texttt{ConvolutionModule}([w_1,...,w_N]),
\end{equation}
where $\texttt{ConvolutionModule}(\cdot)$ is implemented with layers of CNNs; each resulting vector $W_n$ is in the $d_{model}$ dimension (i.e., $W_n\in{\mathbb{R}}^{d_{model}}$), and $d_{model}$ is a predefined model configuration. The geographical information is processed by the positional encoding module to calculate the corresponding vector for each station:
\begin{equation} 
    \label{PositionEmbedding_en}
    [G_1,...,G_N] = \texttt{PositionalEncoding}([g_1,...,g_N]),
\end{equation}
where the calculation function $\texttt{PositionalEncoding}(\cdot)$ is implemented using pairs of sinusoidal functions (i.e., sine and cosine) as usual \cite{vaswani2017attention}, and $G_n$ is also in the $d_{model}$ dimension. Afterward, the two sets of features are summed together with learnable weighting factors for each station individually:
\begin{equation} 
    \label{Before_HiddenState_en}
    \begin{split}
        [H^0_1,\ldots,H^0_N] &= \texttt{ADD}([\alpha_1,...,\alpha_N]\cdot[W_1,...,W_N], \\
        &\quad [(1-\alpha_1),...,(1-\alpha_N)]\cdot[G_1,...,G_N]),
    \end{split}
\end{equation}
where the range of each weighting factor $\alpha_n$ is 0-1. The weighting factors are used to automatically adjust the balance between the waveform and geographical information. Afterward, early locality-specific embeddings $[l^e_1, \ldots, l^e_N]\in{\mathbb{R}}^{d_{model}\times{N}}$ are combined with $[H^0_1,\ldots,H^0_N]$ to compensate for the special characteristics and biases of each station:
\begin{equation} 
    \label{Early_Locality_Specific_en}
    [H^1_1, \ldots, H^1_N] = \texttt{ADD}([H^0_1,\ldots,H^0_N],[l^e_1, \ldots, l^e_N]).
\end{equation}
The set of early locality-specific embeddings are model parameters that are learned as the model is trained.

So far, the encoder is used to process the two modality statistics and compress them into feature representations with auxiliary learnable station-specific hidden characteristics (i.e., early locality-specific embeddings). Next, these representations are fed into the decoder. The feature blending module is first used to process the set of features. To investigate the relationships between stations and simultaneously consider all the stations, the feature blending module is constructed by staking multiple layers of self-attention-based models, such as Transformers or Conformers:
\begin{equation} 
    [H^2_1, \ldots, H^2_N] = \texttt{FeatureBlending}([H^1_1, \ldots, H^1_N]),
\end{equation}
where each $H^2_n$ is a ${d_{model}}$ dimensional vector. Next, a set of late locality-specific embeddings $[l^l_1, \ldots, l^l_N]\in{\mathbb{R}}^{d_{model}\times{N}}$ is applied to inject the station-specific factors again into the representations:
\begin{equation} 
    [H^3_1, \ldots, H^3_N] = \texttt{ADD}([H^2_1, \ldots, H^2_N],[l^l_1, \ldots, l^l_N]).
\end{equation}
Finally, the prediction module is introduced to deliver predictions for locations corresponding to the input stations:
\begin{equation} 
    [D_1, \ldots, D_N] = \texttt{PredictionModule}([H^3_1, \ldots, H^3_N]).
\end{equation}
Two kinds of prediction functions are introduced in Section \ref{Learning}.

\subsection{Learning Objectives}\label{Learning}
As an earthquake early warning system, SENSE is designed to predict intensity levels for target locations. This paper proposes a simple discrete classification method and a continuous counterpart with a mixture density network \cite{Bishop}.

For the former, the prediction module in SENSE is created by a simple feedforward neural network (FFNN) followed by a sigmoid activation function:
\begin{equation} 
    \label{discrete}
    \begin{split}
        [D^{dis}_1, \ldots, D^{dis}_N]&=\texttt{PredictionModule}([H^3_1, \ldots, H^3_N]) \\
        &=\texttt{Sigmoid}(\texttt{FFNN}([H^3_1, \ldots, H^3_N])),
    \end{split}
\end{equation}
where $D^{dis}_n$ is a $C$ dimensional vector (i.e., $D^{dis}_n\in{\mathbb{R}}^{C}$), and $C$ denotes the number of predefined intensity levels. Thus, the result is the probability of each intensity level, and we selected the highest as the final output. The objective of the training is to minimize the classification errors calculated by cross-entropy loss:
\begin{equation} 
    \label{discrete_loss}
    \begin{split}
        \mathcal{L}_{dis}&=-\sum_{n=1}^{N}\sum_{y\leq y_n}\text{log}P(y|D^{dis}_n),
    \end{split}
\end{equation}
where $y_n$ denotes the ground truth label for the $n^{th}$ station. In other words, $y_n$ is the category of the largest PGA actually observed on the $n^{th}$ station. A reasonable scenario is that if the intensity level of the $n^{th}$ station is $y_n$, the low-intensity classes (i.e., $y\leq y_n$) should also be triggered. Thus, the learning objective is to onset all classes that are smaller than the ground truth $y_n$ while offsetting other classes.

\begin{figure}[t]
    \centering
    \includegraphics[width=0.50\textwidth]{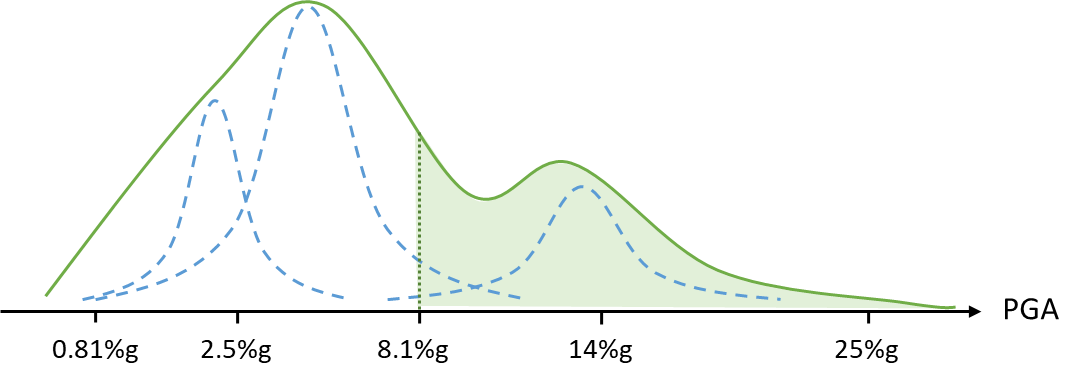}
    \caption{A simple example of using a Gaussian mixture model to approximate the probability density at a given location. The blue dashed lines represent three Gaussians with different means and standard deviations. The green curve denotes the result of mixing the three Gaussians. The green area below the green curve is the probability of the earthquake exceeding 8.1\%g occurring at the location.} 
    \label{fig:Gaussion}
\end{figure}

For the continuous counterpart, we employed a mixture density network to approximate the complex functionality of the PGA behavior. A Gaussian mixture model returned by the mixture density network was used to calculate the probability of each PGA value for a target location. According to the probability, we can decide whether or not to issue an alarm. In sum, the prediction module is a mixture density network implemented by an FFNN in this study, and the output is Gaussian mixture statistics:
\begin{equation} 
    \label{continuous}
    \begin{split}
        [D^{cont}_1, \ldots, D^{cont}_N]&=\texttt{PredictionModule}([H^3_1, \ldots, H^3_N]) \\
        &=\texttt{FFNN}([H^3_1, \ldots, H^3_N]).
    \end{split}
\end{equation}
If the number of Gaussians is set to $K$, $D^{cont}_n$ is a ${3K}$-dimensional vector, which comprises $K$ mixture weights $[a_1,\ldots,a_K]$, $K$ mean values $[\mu_1,\ldots,\mu_K]$, and $K$ standard deviations $[\sigma_1,\ldots,\sigma_K]$. To maintain the property of the mixture weights (i.e., $\sum_{k=1}^{K}a_{k}=1$), a softmax activation function is applied to $[a_1,\ldots,a_K]$. A rectified linear unit activation function (ReLU) is applied to $[\sigma_1,\ldots,\sigma_K]$. The training loss is calculated by minimizing the negative log-likelihood:
\begin{equation} 
    \begin{split}
        \mathcal{L}_{cont}&=-\sum_{n=1}^{N}\text{log}p(y_n|D^{cont}_n) \\
        &=-\sum_{n=1}^{N}\text{log}\left(\sum_{k=1}^{K}a_k\mathcal{N}(y_n|\mu_k,\sigma_k)\right),
    \end{split}
\end{equation}
\begin{equation} 
    \begin{split}
        \mathcal{N}(y_n|\mu_k,\sigma_k)&=\frac{1}{\sigma_k\sqrt{2\pi}}\exp^{-\frac{1}{2}(\frac{y_n-\mu_k}{\sigma_k})^2},
    \end{split}
\end{equation}
where $y_n$ denotes the maximum PGA observed at the $n^{th}$ station. During inference, for a given location, we can calculate the occurring probability for each PGA $u_{PGA}$ by integrating from infinite back to $u_{PGA}$:
\begin{equation} 
    \begin{split}
        P(u_{PGA}|D_n^{cont})=\sum_{k=1}^{K}a_k\left(\int_{u_{PGA}}^{\infty}\mathcal{N}(y|\mu_k,\sigma_k)dy\right).
    \end{split}
\end{equation}
An earthquake alarm would be issued if the probability of a predefined PGA exceeds a threshold. Fig. \ref{fig:Gaussion} illustrates a simple example.

\subsection{Summary}
The proposed SENSE model exhibits an encoder–decoder structure. The encoder is mainly used to process the information collected from seismic stations. For that reason, a convolution module is employed to process the waveform, and a positional encoding module is used to transform physical location information into digital representations. An early locality-specific encoding module is designed to model the detailed characteristics of each station. Based on the meticulous design, the input information can be encapsulated as a set of high-level abstractive feature vectors. After the encoder, a decoder, comprising a feature blending module, late locality-specific encoding module, and prediction module, is introduced. The feature blending module is a vital component of SENSE. It can model the relationships among stations, remove impurities, and keep important information in feature vectors. After the process, a late locality-specific encoding module is further applied to compensate for the unique biases of each station again and generate the resulting representations. Thereafter, the prediction module translates the representations into intensity predictions. 

Seismological information can be categorized into two broad types: waveform and geographical information. Waveform information includes seismic wave data, including amplitude, frequency, wave velocity, and waveform shape. Oppositely, geographical information refers to various surface features of the earth, such as topography, geomorphology, and geological structures \cite{stidham1999three}. SENSE employs the convolution module at the beginning of the encoder to digest the waveform information, and the positional encoding module is employed to encode the raw geographic data. In addition, SENSE introduces early and late locality-specific encodings to enable the model to automatically learn the hidden characteristics of each station to compensate for the input. By combining the waveform, geographical, and hidden information, SENSE establishes a connection between the physical characteristics of the earth and seismic wave data.  Consequently, SENSE is expected to lead to a more comprehensive understanding of seismic phenomena and improve earthquake prediction accuracy \cite{semblat2009waves,bloemheuvel2022multivariate,van2020automated}. 

\begin{figure*}
\begin{subfigure}{.5\textwidth}
  \centering
  \includegraphics[width=0.65\linewidth]{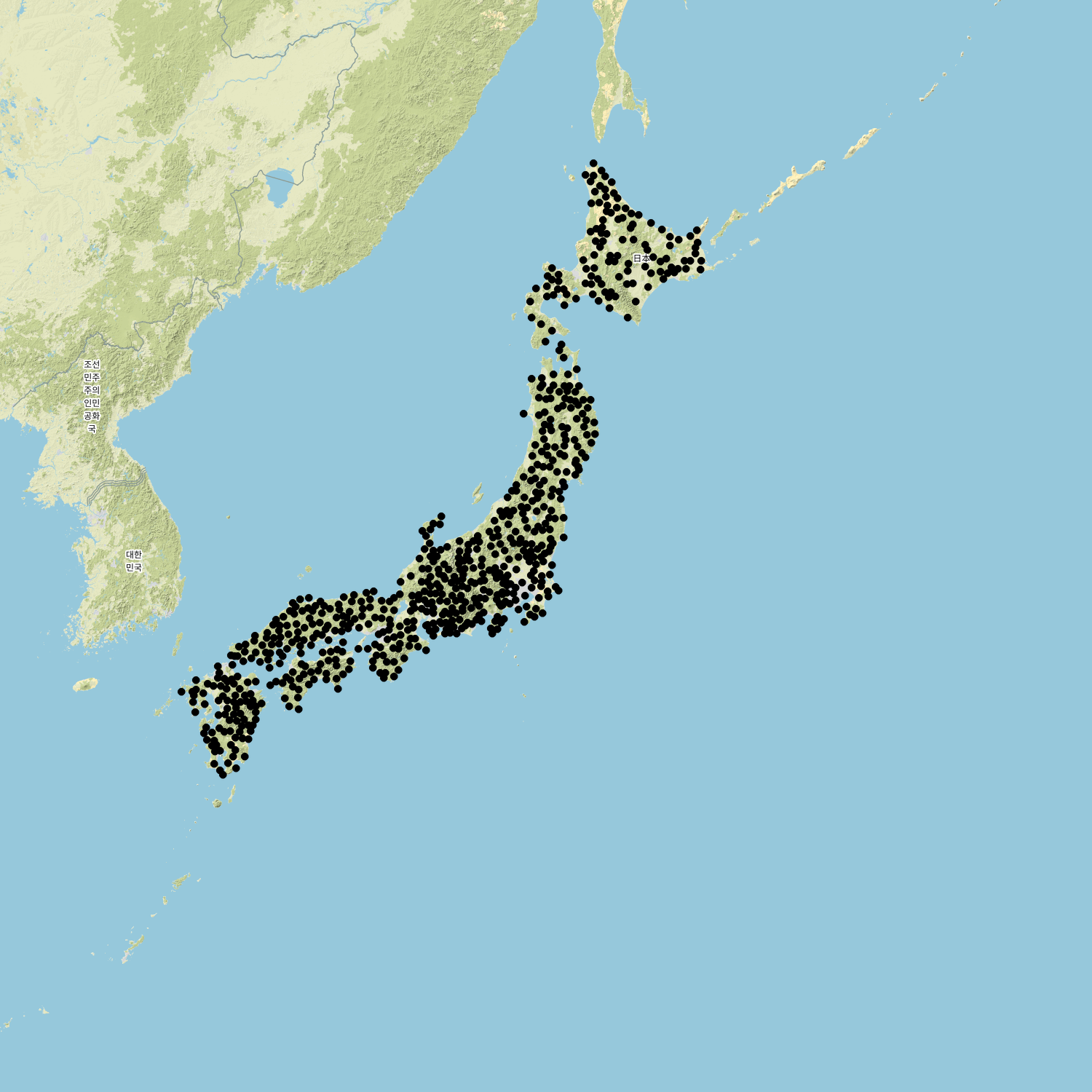}
  \caption{}
  \label{fig:TransformerforJapan}
\end{subfigure}
\begin{subfigure}{.5\textwidth}
  \centering
  \includegraphics[width=0.65\linewidth]{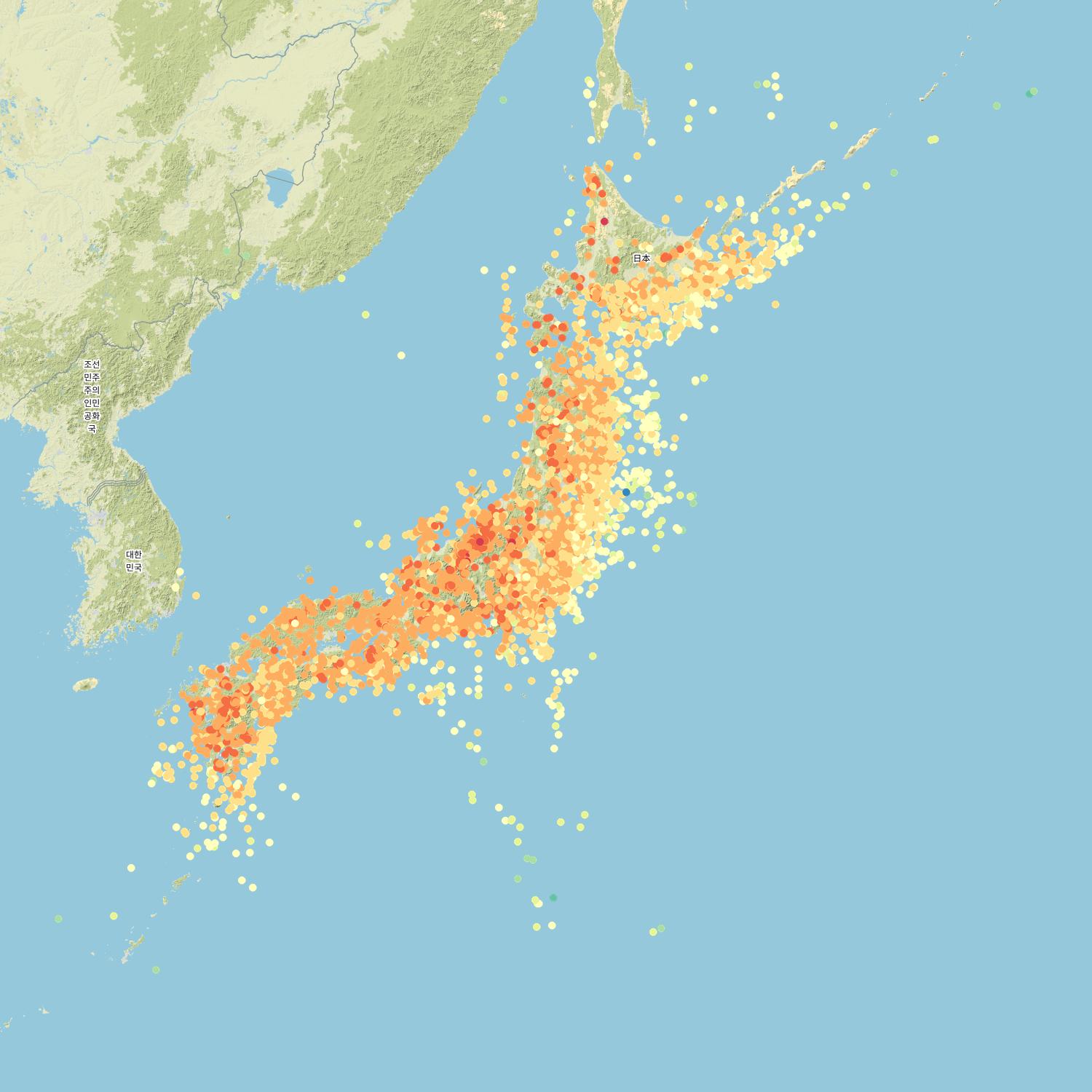}
  \includegraphics[width=0.17\linewidth]{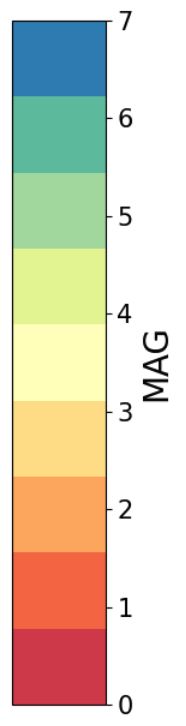}
  \caption{}
  \label{fig:SENSEforJapan}
\end{subfigure}
\caption{Japan dataset. (a) Location distribution of stations. (b) Distribution of earthquake event magnitudes.}
\label{fig:Japan_dataset}
\end{figure*}

\begin{figure*}
\begin{subfigure}{.5\textwidth}
  \centering
  \includegraphics[width=0.65\linewidth]{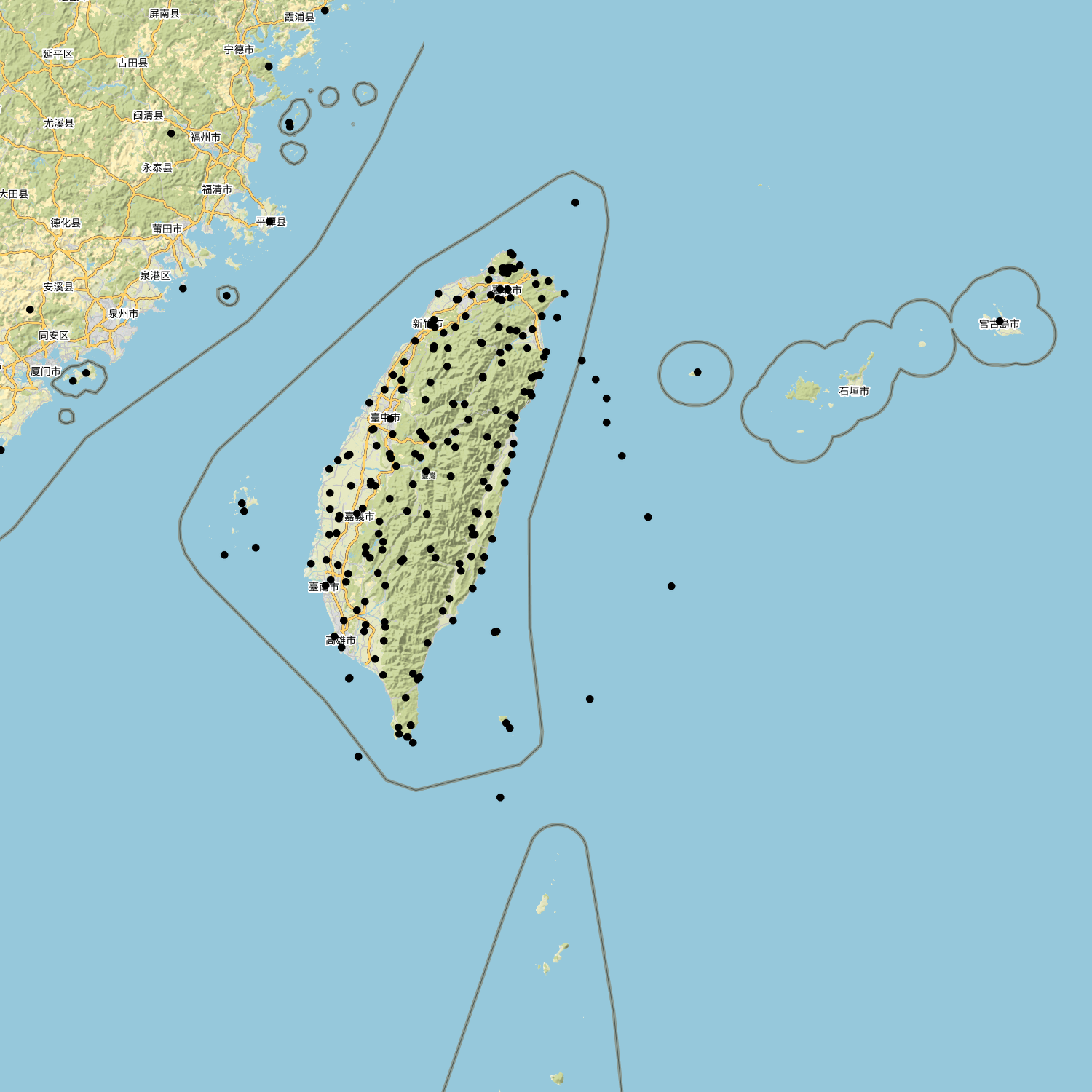}
  \caption{}
  \label{fig:TransformerforTaiwan}
\end{subfigure}
\begin{subfigure}{.5\textwidth}
  \centering
  \includegraphics[width=0.65\linewidth]{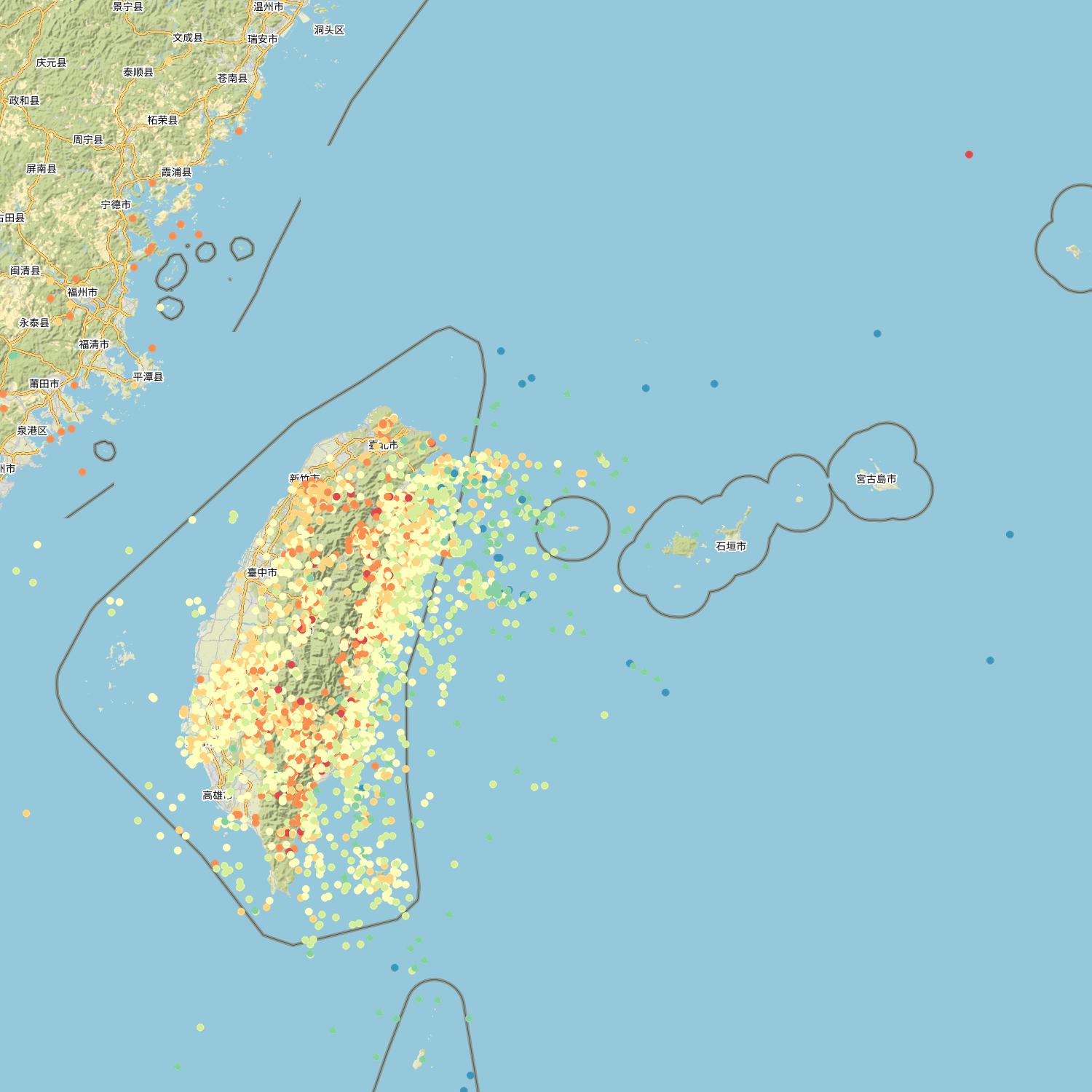}
  \includegraphics[width=0.17\linewidth]{images/my_custom_colorbar.png}
  \caption{}
  \label{fig:SENSEforTaiwan}
\end{subfigure}
\caption{Taiwan Dataset. (a) Location distribution of stations. (b) Distribution of earthquake event magnitudes.}
\label{fig:Taiwan_dataset}
\end{figure*}

\begin{table}[h]
    \small
    \centering
    \caption{Configuration of convolution module used in the SENSE model}
    \label{table:CNN}
    \begin{tabular}{lccc}
        \bottomrule[1pt]
        \multicolumn{1}{c}{Layer} & Number of Filters & Kernel Size & Stride \\
        \hline\hline
        1D Convolution & 8 & 5 & 5 \\
        2D Convolution & 32 & 16$\times$3 & 1$\times$3 \\
        Flatten to 1D &  &   &  \\
        1D Convolution & 64 & 16 & 5,1 \\
        Maximum Pooling &  & 2 & 2 \\
        1D Convolution & 128 & 16 & 1 \\
        Maximum Pooling &  & 2 & 2 \\
        1D Convolution & 32 & 8 & 1 \\
        Maximum Pooling &  & 2 & 2 \\
        1D Convolution & 32 & 8 & 1 \\
        1D Convolution & 16 & 4 & 1 \\
        Flatten to 0D &  &   &  \\
        \toprule[1pt]
    \end{tabular}
\end{table}

\begin{table}[h]
    \small
    \centering
    \caption{Definition of true positive, false positive, true negative, and false negative}
    \label{table:Measures}
    \begin{tabular}{cc}
        \bottomrule[1pt]
        Indicator & Definition \\
        \hline\hline
        \multicolumn{1}{c}{True Positive} & \multicolumn{1}{m{5.5cm}}{The model predicts an alarm that is earlier than the annotated time.}\\
        \hline
        \multicolumn{1}{c}{False Positive} & \multicolumn{1}{m{5.5cm}}{The model predicts an alarm, but there is no need for an alarm.}  \\
        \hline
        \multicolumn{1}{c}{True Negative} & \multicolumn{1}{m{5.5cm}}{The model predicts no alarm, and there is no need for an alarm.}  \\
        \hline
        \multicolumn{1}{c}{False Negative} & \multicolumn{1}{m{5.5cm}}{The model does not issue an alert, but it should be alarmed, or the model predicts an alarm, but it is later than the annotated time.} \\
        \toprule[1pt]
    \end{tabular}
\end{table}

\begin{table*}[]
\renewcommand\arraystretch{1.2}
\caption{Experimental results of the proposed SENSE model on the Japan Dataset}
\label{table:ResultJapan}
\centering
\resizebox{\textwidth}{!}{
\begin{tabular}{ccccccc|cccccc}
\bottomrule[1pt]

\multicolumn{1}{c}{} & 
\multicolumn{6}{c|}{Discrete Classification} & 
\multicolumn{6}{c}{Continuous Estimation} \\ 

\multicolumn{1}{c}{} & 
\multicolumn{3}{c}{Transformer} & 
\multicolumn{3}{c|}{Conformer} & 
\multicolumn{3}{c}{Transformer} & 
\multicolumn{3}{c}{Conformer} \\ 
\cline{2-13}

\multicolumn{1}{c}{PGA[g]} & 
\multicolumn{1}{c}{Precision} & 
\multicolumn{1}{c}{Recall} & 
\multicolumn{1}{c}{F1} & 
\multicolumn{1}{c}{Precision} & 
\multicolumn{1}{c}{Recall} & 
\multicolumn{1}{c|}{F1} & 
\multicolumn{1}{c}{Precision} & 
\multicolumn{1}{c}{Recall} & 
\multicolumn{1}{c}{F1} & 
\multicolumn{1}{c}{Precision} & 
\multicolumn{1}{c}{Recall} & 
\multicolumn{1}{c}{F1} \\ 
\hline
\hline

\multirow{1}{*}{1\%} & 
0.860 & 0.611 & 0.715 & 
0.300 & 0.198 & 0.239 & 
0.647 & 0.779 & 0.707 & 
0.765 & 0.469 & 0.581 \\

\multirow{1}{*}{2\%} & 
0.833 & 0.617 & 0.709 & 
0.411 & 0.196 & 0.265 & 
0.591 & 0.737 & 0.656 & 
0.716 & 0.385 & 0.501 \\

\multirow{1}{*}{5\%} & 
0.818 & 0.611 & 0.699 & 
0.294 & 0.354 & 0.321 & 
0.529 & 0.667 & 0.590 & 
0.627 & 0.250 & 0.385 \\

\multirow{1}{*}{10\%} & 
0.700 & 0.626 & 0.661 & 
0.238 & 0.603 & 0.341 & 
0.447 & 0.648 & 0.549 & 
0.510 & 0.158 & 0.241 \\

\multirow{1}{*}{20\%} & 
0.428 & 0.800 & 0.557 & 
0.651 & 0.410 & 0.503 &
0.495 & 0.470 & 0.482 & 
0.450 & 0.090 & 0.150 \\

\toprule[1pt]
\end{tabular}
}
\end{table*}

\begin{table*}[]
\renewcommand\arraystretch{1.2}
\caption{Experimental results of the proposed SENSE model on the Taiwan Dataset}
\label{table:ResultTaiwan}
\centering
\resizebox{\textwidth}{!}{
\begin{tabular}{ccccccc|cccccc}
\bottomrule[1pt]

\multicolumn{1}{c}{} & 
\multicolumn{6}{c|}{Discrete Classification} & 
\multicolumn{6}{c}{Continuous Estimation} \\ 

\multicolumn{1}{c}{} & 
\multicolumn{3}{c}{Transformer} & 
\multicolumn{3}{c|}{Conformer} & 
\multicolumn{3}{c}{Transformer} & 
\multicolumn{3}{c}{Conformer} \\ 
\cline{2-13}

\multicolumn{1}{c}{PGA[g]} & 
\multicolumn{1}{c}{Precision} & 
\multicolumn{1}{c}{Recall} & 
\multicolumn{1}{c}{F1} & 
\multicolumn{1}{c}{Precision} & 
\multicolumn{1}{c}{Recall} & 
\multicolumn{1}{c|}{F1} & 
\multicolumn{1}{c}{Precision} & 
\multicolumn{1}{c}{Recall} & 
\multicolumn{1}{c}{F1} & 
\multicolumn{1}{c}{Precision} & 
\multicolumn{1}{c}{Recall} & 
\multicolumn{1}{c}{F1} \\ 
\hline
\hline

\multirow{1}{*}{0.81\%} & 
0.627 & 0.273 & 0.380 & 
0.503 & 0.286 & 0.365 & 
0.718 & 0.669 & 0.692 & 
0.576 & 0.694 & 0.629 \\

\multirow{1}{*}{2.5\%} & 
0.494 & 0.377 & 0.428 & 
0.441 & 0.439 & 0.440 & 
0.636 & 0.623 & 0.630 & 
0.518 & 0.494 & 0.506 \\

\multirow{1}{*}{8.1\%} & 
0.459 & 0.537 & 0.495 & 
0.298 & 0.747 & 0.426 & 
0.523 & 0.484 & 0.503 & 
0.316 & 0.453 & 0.372 \\

\multirow{1}{*}{14\%} & 
0.449 & 0.500 & 0.473 & 
0.463 & 0.568 & 0.510 & 
0.426 & 0.523 & 0.469 & 
0.153 & 0.250 & 0.190 \\

\multirow{1}{*}{25\%} & 
0.194 & 0.875 & 0.318 & 
0.333 & 0.375 & 0.353 &
0.300 & 0.375 & 0.333 & 
0.400 & 0.250 & 0.308 \\

\toprule[1pt]
\end{tabular}
}
\end{table*}

\section{EXPERIMENTS}
\subsection{DataSets}
This study used two national-level datasets from highly seismically active regions with dense seismic networks: the Japan and Taiwan datasets. Both datasets were divided into training, validation, and test sets to perform model training, hyperparameter tuning, and fair evaluation. We employed an event-based splitting approach by assigning the records of a specific event to the same subset. The Japan dataset was sourced from Okada and Kasahara \cite{okada2004recent}, and the Taiwan Central Weather Administration provided the Taiwan dataset. To evaluate the model performance, five PGA values were selected for shaking from light to strong. To align with previous studies, [1\%g, 2\%g, 5\%g, 10\%g, 20\%g] and [0.81\%g, 2.5\%g, 8.1\%g, 14\%g, 25\%g] were selected and used for the Japan and Taiwan datasets, respectively \cite{chiang2022neural, Münchmeyer2020b}.

\subsubsection{Japan Dataset}
The Japan dataset comprises 13,512 events compiled from the NIED Kiban Kyoshin network (KiK-net) catalog between 1997 and 2018 \cite{KiK-net}, as shown in Fig. \ref{fig:Japan_dataset}. The data includes records of triggered strong motion, with each trace containing 15 seconds of pre-trigger data for a total length of 120 seconds. Each station had two three-component strong-motion sensors (one at the surface and one underground). The dataset was split into training, validation, and test sets at a 60:10:30 ratio. The training set ended in March 2012, the test set began in August 2013, and the intervening events were used as the validation set. The total number of stations was 707. 

\subsubsection{Taiwan Dataset}
The Taiwan dataset comprises 9,311 earthquake events recorded by the dense seismic network in Taiwan from 2012 to 2021, as shown in Fig. \ref{fig:Taiwan_dataset}. Each event includes seismic waveforms recorded by three-component strong-motion sensors. We partitioned the dataset into training, validation, and test sets in chronological order at a ratio of 60:25:15. The training, validation, and test sets included 2012–2017, 2018–2019, and 2020–2021 events, respectively. There were 250 stations in the dataset.

\subsection{Implementation Details}
\subsubsection{Training Configurations}
The training process was stabilized to obtain better model parameters through three training procedures. In the first round, the parameters were updated, except for the early locality-specific encoding, and the learnable weighting factors $\{\alpha_1,\ldots,\alpha_N\}$ were set to $0.5$. In the second round, the early locality-specific encoding was trained, except for the learnable weighting factors. Finally, the model parameters were simultaneously updated. For the Japan dataset, the training epochs were 100, 40, and 40. For the Taiwan dataset, the training epochs were 50, 20, and 20 for the three steps. Table \ref{table:CNN} lists the detailed configuration of the convolution module. The feature blending module comprises six layers of either the Transformer encoder or Conformer. For the Transformer encoder and Conformer, the hidden dimension of the FFNN was $1000$, and the head number was set to $10$. Considering that five PGA values were selected to report the experimental results, the number of classes $C$ for the discrete classification objective was set to $5$. The $\texttt{FFNN}(\cdot)$ in the prediction module for the discrete case comprised five hidden layers in the sizes of 500, 150, 100, 50, and 30 (cf. (\ref{discrete})). In the continuous case, $\texttt{FFNN}(\cdot)$ was implemented by six hidden layers in the sizes of 500, 150, 100, 50, 30, and 10 (cf. (\ref{continuous})). 

It is important to highlight that the training procedure is carefully crafted based on empirical insights to optimize both model stability and performance. During the first phase, the model focuses on learning general seismic patterns and capturing broad relationships among input features. Concurrently, late locality-specific embeddings, which encapsulate station-specific biases and corrections, are updated to align core feature representations with the unique characteristics of each station. This phase establishes a foundational mapping between general seismic features and localized adjustments, creating a stable baseline and preparing the model for more fine-grained, station-specific tuning. The second phase introduces early locality-specific embeddings, allowing the model to incorporate nuanced, location-specific information. By isolating the training of these embeddings, the model can encode unique features of each seismic station without disrupting the broader relationships learned in the initial phase. This modular approach enables the seamless integration of specific and general information, mitigating the risks of overfitting or instability. In the final phase, all parameters, including core features and both early and late locality-specific embeddings, are updated simultaneously. This comprehensive fine-tuning process harmonizes general seismic insights with station-specific characteristics, ensuring a balanced and accurate predictive capability. By progressively building complexity and refining accuracy across these phases, the model is designed to achieve robustness and high performance as an earthquake early warning system. While alternative training configurations might further enhance results, a comprehensive search for optimal settings was constrained by computational resources.

\subsubsection{Evaluation Metrics}
To evaluate the performance of the proposed SENSE model, we defined the true positives (TP), false positives (FP), false negatives (FN), and true negatives (TN) as described in Table \ref{table:Measures}. Based on these statistics, we computed the precision, recall, and F1 scores:
    \begin{equation} 
        \label{Precision_en}
        \begin{aligned}
            Precision &= \frac{TP}{TP+FP}, \\
        \end{aligned}
    \end{equation}
    \begin{equation} 
        \label{Recall_en}
        \begin{aligned}
            Recall &= \frac{TP}{TP+FN}, \\
        \end{aligned}
    \end{equation}
    \begin{equation} 
        \label{F1-Score_en}
        \begin{aligned}
            F1-score &= \frac{2 \times Precision \times Recall}{Precision+Recall}. \\
        \end{aligned}
    \end{equation}

\subsection{Main Results}
In the first set of experiments, we evaluated the proposed SENSE model using the Japan and Taiwan datasets. Tables \ref{table:ResultJapan} and \ref{table:ResultTaiwan} present the results for different datasets and model configurations. Several observations were drawn from the results. Pairing the discrete objective with the Transformer model for the Japan dataset resulted in better F1 scores in all five PGA levels. In most cases, the precision scores exceeded 0.8, and the recall rates exceeded 0.6. However, the experimental results were poor when the discrete training objective was combined with the Conformer model. For the continuous objective, the behavior of the Transformer and Conformer models differed considerably. The Transformer and Conformer models obtained better recall and precision scores, respectively. The recall scores of the Conformer model were unsatisfactory in the discrete and continuous cases. Summarily, the Transformer model achieved better results with the Japan dataset regardless of whether the discrete or continuous training targets.

Next, we analyzed the results of the Taiwan dataset. For the discrete training objective, the F1 scores of the Transformer and Conformer models were close for each PGA level while the precision and recall scores were mixed. In the continuous case, the Transformer was generally better than the Conformer in terms of precision, recall, and F1 scores. In a nutshell, the first set of experiments showed that pairing the continuous training objective and Transformer model may be a better and more stable choice for SENSE because the setting can lead to acceptable results for both datasets. 

\begin{table*}[h]
    \centering
    \renewcommand\arraystretch{1.2}
    \caption{Experimental results of the ISMP, TEAM, and SENSE model on the Taiwan dataseet} 
    \label{table:ResultISMPTEAM}
    \resizebox{\textwidth}{!}{
        \centering
        \tiny
        
        \begin{tabular}{cccccccccc}
        \bottomrule
            \multicolumn{1}{c}{} & 
            \multicolumn{3}{c}{ISMP} & 
            \multicolumn{3}{c}{TEAM} &
            \multicolumn{3}{c}{SENSE} \\ 
            \cline{2-10}

            \multicolumn{1}{c}{PGA[g]} & 
            \multicolumn{1}{c}{Precision} & 
            \multicolumn{1}{c}{Recall} & 
            \multicolumn{1}{c}{F1} & 
            \multicolumn{1}{c}{Precision} & 
            \multicolumn{1}{c}{Recall} & 
            \multicolumn{1}{c}{F1} & 
            \multicolumn{1}{c}{Precision} & 
            \multicolumn{1}{c}{Recall} & 
            \multicolumn{1}{c}{F1} \\
            \hline\hline
    
            \multicolumn{1}{c}{0.81\%} & 
            \multicolumn{1}{c}{-} & 
            \multicolumn{1}{c}{-} & 
            \multicolumn{1}{c}{-} &
            \multicolumn{1}{c}{0.211} & 
            \multicolumn{1}{c}{0.554} & 
            \multicolumn{1}{c}{0.306} &
            \multicolumn{1}{c}{0.718} & 
            \multicolumn{1}{c}{0.669} & 
            \multicolumn{1}{c}{0.692} \\
    
            \multicolumn{1}{c}{2.5\%} &
            \multicolumn{1}{c}{0.303} & 
            \multicolumn{1}{c}{0.860} & 
            \multicolumn{1}{c}{0.448} &
            \multicolumn{1}{c}{0.188} & 
            \multicolumn{1}{c}{0.425} & 
            \multicolumn{1}{c}{0.261} &
            \multicolumn{1}{c}{0.636} & 
            \multicolumn{1}{c}{0.623} & 
            \multicolumn{1}{c}{0.630} \\
            
            \multicolumn{1}{c}{8.1\%} & 
            \multicolumn{1}{c}{0.135} & 
            \multicolumn{1}{c}{0.874} & 
            \multicolumn{1}{c}{0.233} & 
            \multicolumn{1}{c}{0.111} & 
            \multicolumn{1}{c}{0.326} & 
            \multicolumn{1}{c}{0.165} &
            \multicolumn{1}{c}{0.523} & 
            \multicolumn{1}{c}{0.484} & 
            \multicolumn{1}{c}{0.503} \\
            
            \multicolumn{1}{c}{14\%} & 
            \multicolumn{1}{c}{-} & 
            \multicolumn{1}{c}{-} & 
            \multicolumn{1}{c}{-} & 
            \multicolumn{1}{c}{0.046} & 
            \multicolumn{1}{c}{0.205} & 
            \multicolumn{1}{c}{0.075} & 
            \multicolumn{1}{c}{0.426} & 
            \multicolumn{1}{c}{0.523} & 
            \multicolumn{1}{c}{0.469} \\
    
            \multicolumn{1}{c}{25\%} & 
            \multicolumn{1}{c}{-} & 
            \multicolumn{1}{c}{-} & 
            \multicolumn{1}{c}{-} & 
            \multicolumn{1}{c}{0.004} & 
            \multicolumn{1}{c}{0.125} & 
            \multicolumn{1}{c}{0.008} & 
            \multicolumn{1}{c}{0.300} & 
            \multicolumn{1}{c}{0.375} & 
            \multicolumn{1}{c}{0.333} \\
            
            \toprule
        \end{tabular}
    }
\end{table*}

\begin{table*}[h]
    \centering
    \renewcommand\arraystretch{1.1}
    \caption{Comparison between ISMP, TEAM, and SENSE models} 
    \label{table:ResultModelsTime}
    \resizebox{\textwidth}{!}{
        \centering
        \tiny
        
        \begin{tabular}{ccccccccccc}
        \bottomrule    
            \multicolumn{2}{c}{} & 
            \multicolumn{3}{c}{ISMP} & 
            \multicolumn{3}{c}{TEAM} &
            \multicolumn{3}{c}{SENSE} \\ 
            \cline{3-11}

            \multicolumn{2}{c}{Model Parameters(M)} & 
            \multicolumn{3}{c}{0.6} & 
            \multicolumn{3}{c}{13} & 
            \multicolumn{3}{c}{13} \\

            \multicolumn{2}{c}{Execution Time(s)} & 
            \multicolumn{3}{c}{0.032} & 
            \multicolumn{3}{c}{0.083} & 
            \multicolumn{3}{c}{0.093} \\
            \hline

            \multicolumn{1}{c}{\multirow{6}{*}{Leading Time(s)}} & 
            \multicolumn{1}{c}{PGA[g]} & 
            \multicolumn{1}{c}{Mean} & 
            \multicolumn{1}{c}{Median} & 
            \multicolumn{1}{c}{Max} & 
            \multicolumn{1}{c}{Mean} & 
            \multicolumn{1}{c}{Median} & 
            \multicolumn{1}{c}{Max} & 
            \multicolumn{1}{c}{Mean} & 
            \multicolumn{1}{c}{Median} & 
            \multicolumn{1}{c}{Max} \\
            \cline{3-11}

            \multicolumn{1}{c}{} &
            \multicolumn{1}{c}{0.81\%} & 
            \multicolumn{1}{c}{-} & 
            \multicolumn{1}{c}{-} & 
            \multicolumn{1}{c}{-} &
            \multicolumn{1}{c}{5.21} & 
            \multicolumn{1}{c}{2.91} & 
            \multicolumn{1}{c}{43.81} &
            \multicolumn{1}{c}{4.30} & 
            \multicolumn{1}{c}{2.48} & 
            \multicolumn{1}{c}{26.11} \\
    
            \multicolumn{1}{c}{} &
            \multicolumn{1}{c}{2.5\%} &
            \multicolumn{1}{c}{6.63} & 
            \multicolumn{1}{c}{5.61} & 
            \multicolumn{1}{c}{13.24} &
            \multicolumn{1}{c}{3.49} & 
            \multicolumn{1}{c}{2.24} & 
            \multicolumn{1}{c}{33.92} &
            \multicolumn{1}{c}{2.85} & 
            \multicolumn{1}{c}{1.76} & 
            \multicolumn{1}{c}{13.84} \\
            
            \multicolumn{1}{c}{} &
            \multicolumn{1}{c}{8.1\%} & 
            \multicolumn{1}{c}{5.98} & 
            \multicolumn{1}{c}{5.67} & 
            \multicolumn{1}{c}{8.32} & 
            \multicolumn{1}{c}{2.62} & 
            \multicolumn{1}{c}{1.67} & 
            \multicolumn{1}{c}{9.67} &
            \multicolumn{1}{c}{1.59} & 
            \multicolumn{1}{c}{1.08} & 
            \multicolumn{1}{c}{8.64} \\
            
            \multicolumn{1}{c}{} &
            \multicolumn{1}{c}{14\%} & 
            \multicolumn{1}{c}{-} & 
            \multicolumn{1}{c}{-} & 
            \multicolumn{1}{c}{-} & 
            \multicolumn{1}{c}{2.53} & 
            \multicolumn{1}{c}{2.27} & 
            \multicolumn{1}{c}{6.17} & 
            \multicolumn{1}{c}{0.86} & 
            \multicolumn{1}{c}{0.70} & 
            \multicolumn{1}{c}{2.63} \\
    
            \multicolumn{1}{c}{} &
            \multicolumn{1}{c}{25\%} & 
            \multicolumn{1}{c}{-} & 
            \multicolumn{1}{c}{-} & 
            \multicolumn{1}{c}{-} & 
            \multicolumn{1}{c}{4.11} & 
            \multicolumn{1}{c}{4.11} & 
            \multicolumn{1}{c}{4.11} & 
            \multicolumn{1}{c}{1.07} & 
            \multicolumn{1}{c}{0.77} & 
            \multicolumn{1}{c}{2.11} \\
            
            \toprule
        \end{tabular}
    }
\end{table*}

\begin{table*}[h]
    \centering
    \renewcommand\arraystretch{1.2}
    \caption{Ablation studies on the SENSE model evaluated using F1 scores}
    \label{table:Ablation}
    \resizebox{\textwidth}{!}{
        \centering
        \small
        
        \begin{tabular}{ccccccc|ccccc}
        \bottomrule[1pt]
            \multicolumn{2}{c}{} & 
            \multicolumn{5}{c|}{Japan} & 
            \multicolumn{5}{c}{Taiwan} \\ 
            \cline{3-12}

            \multicolumn{2}{c}{} & 
            \multicolumn{1}{c}{1\%} & 
            \multicolumn{1}{c}{2\%} & 
            \multicolumn{1}{c}{5\%} & 
            \multicolumn{1}{c}{10\%} & 
            \multicolumn{1}{c|}{20\%} & 
            \multicolumn{1}{c}{0.81\%} & 
            \multicolumn{1}{c}{2.5\%} & 
            \multicolumn{1}{c}{8.1\%} & 
            \multicolumn{1}{c}{14\%} &
            \multicolumn{1}{c}{25\%} \\
            \hline\hline  

            \multicolumn{1}{l}{(A)} & 
            \multicolumn{1}{l}{SENSE} & 
            \multicolumn{1}{c}{0.707} & 
            \multicolumn{1}{c}{0.656} & 
            \multicolumn{1}{c}{0.590} &
            \multicolumn{1}{c}{0.549} & 
            \multicolumn{1}{c|}{0.482} & 
            \multicolumn{1}{c}{0.692} &
            \multicolumn{1}{c}{0.630} & 
            \multicolumn{1}{c}{0.503} & 
            \multicolumn{1}{c}{0.469} &
            \multicolumn{1}{c}{0.333} \\
            \hline 

            \multicolumn{1}{l}{(B)} & 
            \multicolumn{1}{l}{\makecell[l]{\qquad-Weighting Factor  \\  \qquad-Early Locality-specific Encoding}} &
            \multicolumn{1}{c}{0.656} & 
            \multicolumn{1}{c}{0.599} & 
            \multicolumn{1}{c}{0.549} &
            \multicolumn{1}{c}{0.442} & 
            \multicolumn{1}{c|}{0.144} & 
            \multicolumn{1}{c}{0.673} &
            \multicolumn{1}{c}{0.614} & 
            \multicolumn{1}{c}{0.486} &
            \multicolumn{1}{c}{0.293} &
            \multicolumn{1}{c}{0.133} \\
            \hline 

            \multicolumn{1}{l}{(C)} &
            \multicolumn{1}{l}{\makecell[l]{\qquad-Weighting Factor  \\  \qquad-Late Locality-specific Encoding}} & 
            \multicolumn{1}{c}{0.674} & 
            \multicolumn{1}{c}{0.608} & 
            \multicolumn{1}{c}{0.534} & 
            \multicolumn{1}{c}{0.472} & 
            \multicolumn{1}{c|}{0.355} & 
            \multicolumn{1}{c}{0.677} &
            \multicolumn{1}{c}{0.584} & 
            \multicolumn{1}{c}{0.469} & 
            \multicolumn{1}{c}{0.366} &
            \multicolumn{1}{c}{0.364} \\
            \hline 

            \multicolumn{1}{l}{(D)} &
            \multicolumn{1}{l}{\makecell[l]{\qquad-Early Locality-specific Encoding  \\  \qquad-Late Locality-specific Encoding}} & 
            \multicolumn{1}{c}{0.695} & 
            \multicolumn{1}{c}{0.635} & 
            \multicolumn{1}{c}{0.589} & 
            \multicolumn{1}{c}{0.535} & 
            \multicolumn{1}{c|}{0.445} & 
            \multicolumn{1}{c}{0.692} & 
            \multicolumn{1}{c}{0.624} & 
            \multicolumn{1}{c}{0.529} & 
            \multicolumn{1}{c}{0.422} &
            \multicolumn{1}{c}{0.286} \\
            \hline 

            \multicolumn{1}{l}{(E)} &
            \multicolumn{1}{l}{\makecell[l]{\qquad-Weighting Factor  \\ \qquad-Early Locality-specific Encoding  \\ \qquad-Late Locality-specific Encoding}} & 
            \multicolumn{1}{c}{0.664} & 
            \multicolumn{1}{c}{0.588} & 
            \multicolumn{1}{c}{0.506} & 
            \multicolumn{1}{c}{0.388} & 
            \multicolumn{1}{c|}{0.240} & 
            \multicolumn{1}{c}{0.666} & 
            \multicolumn{1}{c}{0.628} & 
            \multicolumn{1}{c}{0.520} & 
            \multicolumn{1}{c}{0.364} &
            \multicolumn{1}{c}{0.200} \\
            
            \toprule[1pt]
        \end{tabular}
    }
\end{table*}

\subsection{Compared to Other Advanced Deep Learning-based Methods}
In the second set of experiments, we compared SENSE with two advanced deep learning-based baselines: the intelligent strong-motion prediction (ISMP) model \cite{chiang2022neural} and the Transformer earthquake alerting model (TEAM) \cite{munchmeyer2021transformer}. The ISMP model is designed to predict early ground motion after an earthquake. The model uses a CNN to extract key features from the initial P-waves and predicts if the PGA of subsequent waves will exceed 80 Gal. The TEAM considers information from multiple seismic stations, as opposed to traditional single-station models. The TEAM can issue warnings for an arbitrary number of locations by inputting waveform information and coordinates from multiple stations, thereby flexibly adapting to various earthquake monitoring networks and warning objectives. TEAM is a Transformer-based model, and the training objective is similar to the continuous case for SENSE. Thus, we compared the TEAM with the SENSE with similar settings, i.e., the continuous training objective and Transformer model. The Taiwan dataset was used as an example, and Table \ref{table:ResultISMPTEAM} presents the results.

The experiments yielded valuable findings. ISMP is a single-station method; therefore, it cannot make predictions for stations that do not receive waveform signals. Thus, the results were only computed based on stations with earthquake records. Contrarily, both TEAM and SENSE considered all the stations. In a way, the comparison is unfair and more critical for TEAM and SENSE. However, Table \ref{table:ResultISMPTEAM} shows that SENSE achieved the best results in all the cases. Second, the performance gap between TEAM and SENSE was wide; thus, the second set of experiments confirmed that SENSE was a better modeling strategy than TEAM in the use of multistation data. Third, SENSE significantly outperformed the TEAM and exceeded ISMP as expected because multistation information is more informative than single-station data. Although SENSE and TEAM are multistation models, different model architectures may yield different results. A possible flaw of TEAM is its decoder design. Although TEAM constitutes a flexible method that can predict the intensity of a target location only by providing the coordinates of the target location, the input in the decoder (i.e., coordinates) considerably differed from the encoded seismic information in the Transformer layer in the encoder. Thus, cross-attention between heterogeneous information may not yield satisfactory and stable results.

In addition to the performance evaluation, we examined the model properties and leading time. Table \ref{table:ResultModelsTime} lists the experimental results. Regarding the model parameters, ISMP was a relatively smaller model than TEAM and SENSE, and TEAM and SENSE had similar sizes. Regarding execution time, all the models were executed on a GeForce GTX TITAN graphics card for testing. ISMP is the fastest because it is a single-station method. TEAM and SENSE were approximately three times slower than ISMP, and the difference between TEAM and SENSE was tolerable. It was complicated to determine the best and worst models in terms of the leading time. However, we analyzed the results of the leading time against their performance listed in Table \ref{table:ResultISMPTEAM}. By doing so, we observed that although ISMP may have obtained better leading times in the mean and median cases, its performances were the worst. In comparison, the proposed SENSE model may be a better choice because although its leading time was a little slower than that of TEAM, its prediction results were significantly better than those of TEAM.

\subsection{Ablation Study}
Next, we examined the efficiency of the components in the SENSE model. Table \ref{table:Ablation} presents the experimental results obtained using the Japan and Taiwan datasets. (A) denotes the SENSE model with the Transformer model, and (B), (C), (D), and (E) are models without certain components. SENSE achieved the best results in most of the cases, indicating that none of the components was dispensable. Second, when comparing (D) and (E), we concluded that the weighting factors, which are used to balance the information between the waveform and geographical statistics (cf. (3)), are important because the performance gaps are to be reckoned with. Third, when comparing (C) and (E), the results showed that early locality-specific encoding provided consistent improvements for the Japan dataset in all the cases and achieved better results for the Taiwan dataset in certain cases. Comparisons between (B) and (E) showed that late locality-specific encoding was more useful in the Japan dataset than in the Taiwan dataset. Considering (B), (C), and (E), we can conclude that early locality-specific encoding appeared to favor large PGA cases, whereas late locality-specific encoding influenced small PGA cases.

\section{Conclusion}
This paper proposes an earthquake early warning system called SENSE. SENSE comprises an encoder and a decoder. The encoder distills and refines statistics from stations. The meticulous design is the convolution module and early locality-specific encoding components. The former employs a stack of CNN layers to manipulate local information embedded in the raw waveform. The latter is a set of automatically learned parameters for each station to store station-dependent characteristics. The special designs of the decoder include the feature blending module, late locality-specific encoding, and prediction module. The feature blending module is built using either the Transformer or Conformer models, both based on the self-attention mechanism. The late locality-specific encoding is a set of parameters for each station to represent station-dependent biases. For the prediction module, we present discrete and continuous learning objectives. Based on a series of experiments, we confirmed the effectiveness of SENSE for improving earthquake early warning systems. The study findings contribute to the reduction in losses caused by earthquakes, enhancement of public safety awareness, and the effective implementation of earthquake disaster management. Future studies will focus on further optimizing the model architecture, enhancing the prediction performance, and reducing the training cost.

\section*{Acknowledgment}
This work was supported by the National Science and Technology Council of Taiwan under Grants NSTC 112-2636-E-011-002 and NSTC 112-2628-E-011-008-MY3 and by the Central Weather Administration of Taiwan. We thank the National Center for High-performance Computing of the National Applied Research Laboratories (NARLabs) in Taiwan for providing computational and storage resources.

\clearpage
\nocite{*}
\bibliography{bibtex/bib/IEEEexample}

\end{document}